\documentclass[sigconf,screen]{acmart}

\usepackage[inline]{enumitem}
\usepackage[acronym]{glossaries}

\makeglossaries

\newacronym{hri}{HRI}{human-robot interaction}
\newcommand{\term}[1]{\textit{#1}}
\newcommand{\species}[1]{\textit{#1}}
\newcommand{\property}[1]{\textbf{#1}}
\newcommand{\expression}[1]{\textsl{#1}}

\copyrightyear{2023}
\acmYear{2023}
\setcopyright{rightsretained}
\acmConference[HRI '23]{Proceedings of the 2023 ACM/IEEE International Conference on Human-Robot Interaction}{March 13--16, 2023}{Stockholm, Sweden}
\acmBooktitle{Proceedings of the 2023 ACM/IEEE International Conference on Human-Robot Interaction (HRI '23), March 13--16, 2023, Stockholm, Sweden}
\acmDOI{10.1145/3568162.3578631}
\acmISBN{978-1-4503-9964-7/23/03}

\makeatletter
\gdef\@copyrightpermission{
\begin{minipage}{0.3\columnwidth}
\href{https://creativecommons.org/licenses/by/4.0/}{\includegraphics[width=0.90\textwidth]{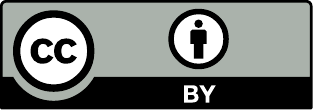}}
\end{minipage}\hfill
\begin{minipage}{0.7\columnwidth}
\href{https://creativecommons.org/licenses/by/4.0/}{This work is licensed under a Creative Commons Attribution International 4.0 License.}
\end{minipage}
\vspace{5pt}
}
\makeatother

\begin{document}

\title{Communicative Robot Signals: Presenting a New Typology for Human-Robot Interaction}

\author{Patrick Holthaus}
\email{p.holthaus@herts.ac.uk}
\orcid{0000-0001-8450-9362}
\affiliation{%
  \institution{University of Hertfordshire}
  \department{School of Physics, Engineering and Computer Science}
  \streetaddress{College Lane}
  \city{Hatfield}
  \country{United Kingdom}
  \postcode{AL10 9AB}
}

\author{Trenton Schulz}
\email{trenton@nr.no}
\orcid{0000-0001-6217-758X}
\affiliation{%
  \institution{Norwegian Computing Center}
  \department{Department of Applied Research in ICT}
  \city{Oslo}
  \country{Norway}
}

\author{Gabriella Lakatos}
\email{g.lakatos@herts.ac.uk}
\orcid{0000-0003-1436-7324}
\affiliation{%
  \institution{University of Hertfordshire}
  \department{School of Physics, Engineering and Computer Science}
  \streetaddress{College Lane}
  \city{Hatfield}
  \country{United Kingdom}
  \postcode{AL10 9AB}
}

\author{Rebekka Soma}
\email{rebsaurus@ifi.uio.no}
\orcid{0000-0002-1074-0366}
\affiliation{%
  \institution{University of Oslo}
  \department{Department of Informatics}
  \city{Oslo}
  \country{Norway}
}

\renewcommand{\shortauthors}{Patrick Holthaus, Trenton Schulz, Gabriella Lakatos, \& Rebekka Soma}

\begin{abstract}
We present a new typology for classifying signals from robots when they communicate with humans. For inspiration, we use ethology, the study of animal behaviour and previous efforts from literature as guides in defining the typology. The typology is based on communicative signals that consist of five properties: the origin where the signal comes from, the deliberateness of the signal, the signal's reference, the genuineness of the signal, and its clarity (i.e., how implicit or explicit it is). Using the accompanying worksheet, the typology is straightforward to use to examine communicative signals from previous human-robot interactions and provides guidance for designers to use the typology when designing new robot behaviours.
\end{abstract}




\maketitle

\section{Introduction and motivation}
\label{sec:introduction}

We have all struggled with making ourselves understood.
Communicating in a clear and unambiguous way is a difficult task regardless if the parties involved are human, animal, or robot. Modelling communicative behaviour can aid in implementing communication between humans and robots. We argue that designing communication in \gls{hri} can benefit from a more structured approach to modelling communicative signals since the robot's interactive capabilities are human-made. Such a model or typology can offer consistent and accessible ways to design robot interactions and identify and describe communicated content. It thereby helps to address the following problems:
\begin{enumerate*}[label=(\textit{\alph*})]
    \item designing robot behaviours that communicate what was intended and
    \item examining (mis-)communication in \gls{hri} (and experiments).
\end{enumerate*}
There have been attempts at creating typologies in the past. Some have borrowed ideas from animal communication~\citep{Hegel2011Typology}. Others have based themselves on concepts such as \term{speech acts} and intentional and consequential sounds~\citep{Schulz2021MovementActs}. Others have used psychological inferences and notation from dance to analyze and interpret behaviours~\citep{bianchiniBehavioralObjectsTwofold2016}. Still, none of these attempts are completely satisfying; for example, there is uncertainty around interaction that is implicit by design or implicit by chance.

In this article, we present a new typology to make designing and classifying robot behaviours more straightforward. The theoretical frame for this new typology is rooted in communication theories with a focus on ethology, the field of animal behaviour, and draws inspiration from earlier approaches in characterising communication between humans and robots. The accompanying worksheet for the typology aims to provide roboticists with a better overview and understanding of robot communicative signals by taking a communicative perspective on robot behaviours, addressing one of the central challenges in social robotics~\cite{Sheridan1997challenge}.
We begin by looking at ethology and how an ethological approach can be useful in \gls{hri} (Sect.~\ref{sec:ethology}). We then discuss how ethological and other communication theories are currently being applied in \gls{hri} contexts (Sect.~\ref{sec:applying}). This leads to the presentation of the new typology with examples, illustrating how it can be used in robot behaviour design or applied to existing studies (Sect.~\ref{sec:typology}). We briefly discuss how the typology can complement other approaches and some current limitations with the typology (Sect.~\ref{sec:discussion}) before we conclude the paper (Sect.~\ref{sec:conclusion}).

\section{An ethological approach}
\label{sec:ethology}
Communication between humans and robots can be understood by looking at it from an evolutionary perspective~\cite{Smith2003AnimalSignals,Laidre2013AnimalSignals}. Ethology helps us to find a feasible way of explaining and examining the quality of signals by
\begin{enumerate*}[label=(\textit{\alph*})]
\item looking at how communicative signals evolved and are used in the animal kingdom, and
\item by looking at asymmetric communication between humans and animals and drawing parallels to asymmetric communication between humans and robots
\end{enumerate*}.
Let us examine concepts and theories commonly employed in ethology and see how they can be useful for \gls{hri} and thus for the typology of communicative robot signals we introduce in Sect.~\ref{sec:typology}.

\subsection{Communication definitions}
\label{sec:ethology:com-defs}

Communication in ethology is defined as the behavioural act an animal performs to change the probability another animal will behave in a certain way~\citep{Slater1985Ethology}. Often, this act has adaptive value, at least for the \term{sender} animal. In addition, signals should have an evolutionary history and should be evolved specifically for a communicative function.

Communication, in general, is the process or act of transmitting a message from a sender to a receiver through a channel with the interference of noise~\citep{DeVito1986Handbook}. Some broaden the definition to include that the message transmission is intentional and conveys meaning to bring about change. Others restrict the definition to
\begin{enumerate*}[label=(\textit{\alph*})]
  \item the receiver must respond to the message and
  \item the interacting individuals must be members of the same species~\citep{goodenoughPerspectivesAnimalBehavior2009}.
\end{enumerate*}
A communicational act includes at least two individuals, a \term{sender} and a \term{receiver}. The sender emits a signal that informs the receiver about the sender's inner state or about elements of the environment. The signal's function in the receiver's mental model can change by ontogenetic learning and, thus, one signal can have different functions in different contexts. In this sense, the message of a signal is the function the signal fills in the receiver's mental model.

\Citeauthor{csanyi1988can} split communication based on how well the mental models of the participants correspond~\cite{csanyi1988can}. When a component of the sender's mental model (the signal) has the same function in the receiver's mental model, the correspondence between the two mental models is 100\%. This is  \term{type I communication}. This type of communication rarely occurs, even among individuals of the same species. In \term{type II communication}, the correspondence is below 100\% and cannot be exactly determined. In the inter-species (interspecific) situation, one can only talk about type II communication.
The main challenge of such is that the sender's and receiver's mental models require a common set of signals to communicate.

Based on the signals' nature, researchers differentiate between \term{referential} and \term{non-referential} (motivational) signals~\cite{Kampis1991Systems}. Non-referential signals contain information about the actual motivational state of the animals and are independent of the quality of any external stimuli. Referential signals refer to certain elements of the environment or environmental events independent of the inner state of the sender. Referential communication assumes the signals referring to the elements of the environment have the same role as words in human language, but theoretical and experimental research has shown that function and mechanism cannot be separated since the inner state of the animals cannot be determined. For this reason, researchers developed  \term{functional reference}~\citep{evansMeaningAlarmCalls1993, macedoniaEssayContemporaryIssues1993, evansReferentialSignals1997}. This concept implies that communicational signals have the same function in the animals' communicational system as words in the human language, but it is not assumed that there are similar mechanisms behind the signal production and signal utilisation~\citep{evansReferentialSignals1997}.
Some of these definitions reference interspecific communication, which is relatively rare among animals. The function of communication (both in intra- and interspecific cases) can be to share information or manipulate others.
The most commonly investigated examples of interspecific communication are alarm calls, which are classic examples of functionally referential signals.
For example, Vervet monkeys (\species{Cercopithecus aethiops}) can comprehend some signals of a starling species (\species{Spreo superbus}) living close to them~\citep{cheneyHowMonkeysSee1990}. The starling produces specific alarm calls for different predators (terrestrial or air). Vervet monkeys show adequate reactions to a call, avoiding the type of predator that it refers to. Hence, it can be supposed that the monkeys can comprehend the meaning of these calls. It is, however, debatable if such communication is completely bi-directional or simply eavesdropping from the receiver's side.

An extraordinary case of interspecific communication is when a communicational event develops between an animal species and humans. Greater honeyguides (\species{Indicator indicator}) living in East-Africa have assisted humans living close to them in finding honey for about 20,000 years, guiding them to the hives of wild bees (\species{Apis mellifera}), which are well-protected and out of the honeyguide's reach. The bird's vocal signals and the height of their flight predict the distance to the bees' nest. In addition, communication between the honeyguide and humans is not unidirectional; members of the Boran tribe also call the honeyguide using specific vocal signals~\citep{isackHoneyguidesHoneyGatherers1989} using a dedicated whistle.
The honeyguide responds by flying towards the campsite of the Boran, calls for the human's attention and then guides them to the hive. The bird uses specific vocal signals and movements, knowing that when the Borans obtain the honey, they always leave some honey behind for the honeyguide.
%

\subsection{How can ethology help robotics?}
\label{sec:ethology:robotics}

The etho-robotic approach suggests that \gls{hri} should be considered a specific form of inter-specific interaction. Applying animal models in robot behaviour design thus provides an important alternative in the development of social robots~\citep{miklosiUtilizationSocialAnimals2012}. Human-animal interaction is a valid model for \gls{hri} as both are asymmetric, may start at any age, are much simpler than human-human interactions, and can develop using only nonverbal communicational behaviour on the animal (or robot) side~\citep{lakatosEmotionAttributionNonHumanoid2014}. Hence, much like in type II communication between animals, a common set of communicative signals is required for successful interaction between humans and robots.

Dogs provide an excellent biological model since they adapted to the human social environment exceptionally well, developing specific interspecific communicational skills towards humans, enabling them to participate in numerous complex social interactions with us on a daily basis~\citep{topalChapterDogModel2009}. Dogs have not only evolved exceptional interspecific communicational skills but also develop a strong attachment bond with humans, making them life-long companions~\citep{topalAttachmentBehaviorDogs1998}. This approach argues that the implementation of dog-analogue behaviours in social robots could lead to more believable and acceptable robotic companions~\citep{miklosiUtilizationSocialAnimals2012, miklosiEthoroboticsNewApproach2017a, vinczeEthologicallyInspiredHumanrobot2012}.

\section{Communication Theories and HRI}
\label{sec:applying}

Let us examine some \gls{hri} studies that have used behaviour drawn
from animals, humans, and machines for specific interactions. Then, we
present how general theory and the arts can also be used for designing
communication. We also present some earlier attempts at
characterising communicative robot signals in general.

\subsection{Studies inspired by etho-robotics, human, and artificial behaviour}
\label{sec:applying:etho-robotics-studies}


There are several experimental studies that have used an etho-robotics approach, and several have been inspired by behaviours from dogs. One study used dogs' motivational (non-referential) signals as a model to design emotionally expressive behaviour (happiness, fear, and guilt) for robots and investigated whether participants could recognise the  emotions expressed by the robot and interact with it accordingly~\citep{lakatosEmotionAttributionNonHumanoid2014}. The results suggested people attribute emotions to a social robot and interact with it according to the expressed emotional behaviour.  A recent study designed guidelines for 11 affective expressions for the Miro robot and applied  behaviour design inspired by dog behaviour among others, and evaluated the expressions through an online video study. All expressions were recognised significantly above the chance level~\citep{ghafurianZoomorphicMiroRobot2022}. So, the dog-inspired behaviour proved a suitable medium for making people attribute emotional states to a non-humanoid robot.

Another study used the behaviour of specially trained hearing dogs to design behaviour for an assistive robot.   In the study, the robot successfully led participants to two sound sources using  visual communication signals inspired by the referential behaviour of the hearing dogs. The findings suggested that participants could  correctly interpret the robot's intentions~\citep{koayHeyThereSomeone2013}. This study provided further evidence for dog-inspired social behaviour being a suitable medium for communicating with human participants.

Designing robot behaviours using biologically-inspired behaviour is not restricted to only dog behaviour. A recent study used a Roomba and a LED strip to imitate aquatic animals' use of light, examining how bio-luminescence, colour science, and
colour psychology can be used in \gls{hri}, with a focus on
appearing attractive or
hostile~\citep{songBioluminescenceInspiredHumanRobotInteraction2018}.
The study
found that low-intensity blue lights in a circle or split pattern were
more attractive to participants, while high-intensity red lights in a
blinking or breathing pattern were perceived as
hostile~\citep{songBioluminescenceInspiredHumanRobotInteraction2018}.
Note that aquatic animals' communicative signals are \emph{not} used in interspecific interactions with humans and hence, have no biological relevance to human communication. This suggests that aquatic animals may not provide a suitable model for robot behaviour design in \gls{hri}, and explanations for these findings may be more complex even though the signals were perceived as expected by the participants.


Human behaviour can be a source for creating robot signals, but the result may not always be perceived as human-like.
One study used a Nao with three presenting-behaviours:
\begin{enumerate*}[label=(\textit{\alph*})]
  \item \term{leader} where gaze and nod were not derived from human response (machine-like), 
  \item \term{follower} version where the robot follows the person's gaze and nodding behaviour (machine-like), and
  \item \term{semi-follower} where the robot would sometimes follow the person and sometimes try to lead (human-like)~\cite{thepsoonthornDoesRobotHumanbased2017}.
\end{enumerate*}
Participants watched and were asked to pick the most human-like behaviour. The follower and semi-follower behaviour were picked as more human-like over the leader behaviour, but some participants preferred the nodding behaviour of the follower behaviour, while others preferred the gaze and face-tracking in the semi-follower behaviour~\cite{thepsoonthornDoesRobotHumanbased2017}. The study showed that, given a certain context, pursuing human-likeness may not be the most important aspect to design a robot's behaviour.

Human-like behaviour carries the danger that the robot may appear more socially competent than it is.
In some situations, it may be better the robot behaves in a artificial, machine-like manner.
For example, one study~\citep{galloExploringMachinelikeBehaviors2022} had a robot with two behaviours to wait for an elevator in the lobby of a building along with people. In the ``machine-like'' behaviour, the robot waited at a designated spot and only entered the elevator after everyone else had a spot. For the ``natural'' behaviour, the robot joined the rear of the cluster of waiting people and used a (weak) first come, first served policy that was similar to what was observed when humans wait for elevators.
People in the study felt the robot's ``machine-like'' behaviour caused less confusion than the robot's ``natural,'' human-like behaviour~\citep{galloExploringMachinelikeBehaviors2022}.

\subsection{Describing human-robot communication}
\label{sec:applying:characterising}

These examples in Section~\ref{sec:applying:etho-robotics-studies} show robot behaviours can
draw inspiration from animal, human, or machine behaviour. If we use some theory of communication or perhaps adopt methods from the arts, we may design clearer and more understandable signals for communication or notice potential problems.

For example, the concept of \term{speech acts}~\citep{sep-speech-acts} and the concept of intentional and consequential sounds~\citep{knepperImplicitCommunicationJoint2017} was applied to suggest a robot's movements are functional and communicative at the same time (\term{movement acts}~\citep{Schulz2021MovementActs}).
In addition, the communicative part of the movement has an intentional (explicit) component, i.e. designed by the roboticist, and a consequential (implicit) component, i.e. how the person observing a robot's motion interprets that motion~\citep{Schulz2021MovementActs}.
The explicit and implicit components are split between the roboticists and the people interacting with the
robot respectively.
For example, a robot's motion could be designed to be purely functional, but an
observer can implicitly draw unintended meanings out of it (e.g. a Fetch robot's
movement to reset its navigation stack leads to people interpreting
its movement as confusion or being extra
careful).

Designers have deliberately used explicit and implicit elements in the movement of a robotic ottoman to indicate it wanted people to interact with it~\citep{sirkinMechanicalOttomanHow2015}. In another study, the researchers used artistic techniques from improvisational acting to have the robot ottoman's motion interpreted by observers as having different levels of dominance~\citep{liCommunicatingDominanceNonanthropomorphic2019}.
Designers' attempts may not always be successful and a designed behaviour can be misinterpreted, (e.g.  a Cozmo robot's behaviour indicating it was done waiting for a fist bump was misinterpreted as
the robot being sad for other reasons~\citep{pelikanAreYouSad2020}).
It is ultimately
the person interacting with the robot who interprets the
implicitly sent message, but there seems to be a split between an \emph{intended}
implicit signal versus an \emph{unintended} implicit signal.

Turning to the arts, such as
Laban notation for dance~\cite{laviersStyleBasedRoboticMotion2014} and animation principles and techniques~\citep{schulzAnimationTechniquesHumanRobot2019,takayamaExpressingThoughtImproving2011}, can provide inspiration for an intentional, implicit interaction.
\Citet{bianchiniBehavioralObjectsTwofold2016} used Laban notation as inspiration for creating a notation system for non-anthropomorphic robot movement, focusing on the \term{qualia} of gesture: from movement to gesture, and from gesture to behaviour. They noted it was difficult to determine what is or is not a gesture: a gesture cannot be separated from its ``… relation to the postural, dynamic, and contextual organisation of the body''~\citep[p.~16]{bianchiniBehavioralObjectsTwofold2016}.

Movement acts are valid, important classifications that contribute to the understanding
and design of robot communication with humans.
Using methods from the arts, such as animation and dance, can be a good way to implement some of these signals.
But there is a gap in distinguishing what is intentionally implicit (that is, created by the designer to be understood implicitly) and unintentionally implicit (that is, not created intentionally by the designer, but still communicates something to the human).
This split needs to be better developed and presented to minimise misinterpretation.

There have been attempts at describing how communication between humans and robots works. Much of the foundation is from two models. This first is \citet{shannon1948mathematical} that creates a probabilistic communication model that can model signal noise and modulate emphasis in communication. The second is \citet{watzlawick1967pragmatics} and the concept of \term{non-action} summarised succinctly as ``… no matter how one might try, one cannot \emph{not} communicate. Activity or inactivity, words or silence all have message value''~\citep[p.~30]{watzlawick1967pragmatics}.
One model by \citet{bonarini2020communication} examines the channels of hearing, sight, and touch that are available to humans for communicating with robots and suggests to use machine-learning techniques to teach robot behaviours and then benchmark their performance~\citep{bonarini2020communication}. This can provide a starting point to explore communication channels.

A more in-depth review of communication theories in \gls{hri} argued that an asymmetrical model for communication is a better fit for human-robot communication since robots are not human and interact with the world differently than humans~\citep{frijns2021communication}, which is compatible with the notion of type~II communication (c.f. Sect.~\ref{sec:ethology:com-defs}).
The review presents the AMODAL-HRI model, which focuses on actions to build a common ground  between a human and a robot~\citep{frijns2021communication}. It also includes the processes in the human and robot for interpreting and acting on the situation~\citep{frijns2021communication}. The model includes guidelines based on Norman's design principles for interaction design~\citep{normanDesignEverydayThings2002}.

Another way to describe the communication is to create a typology for communication for \gls{hri}. Similar to our approach, the typology created by \citet{Hegel2011Typology} was strongly inspired by communication theories between animals. \term{Signs} are the basic term for an entity that transmits information to another entity~\citep{Hegel2011Typology}.
Signs could be divided into \term{signals},  the signs acted on to alter the behaviour of another organism, and \term{cues}, any feature of the world that a receiver could use as a guide to get information about the signaller~\citep{Hegel2011Typology}.
These signals and cues could be further classified as human—resembling human behaviours and features—or artificial—non-human-like signs (e.g. blinking lights or jerky movement) that could communicate information~\citep{Hegel2011Typology}.

In their typology, \citeauthor{Hegel2011Typology} emphasised that signals are \emph{explicitly} designed by the roboticist, while cues may be explicitly or implicitly interpreted by the receiver. That is, there is a distinction between the active nature of a signal and the passive (evolved) nature of a cue~\cite{Hegel2011Typology}. For example,  the active nature of deliberately sending a signal by an animal has a cost (e.g. a predator may also see the signal), and the passive nature of \emph{cues} are usually evolved features like colour or patterns that can indicate toxicity.

We argue that distinguishing between signals and cues in the design of robots adds complexity that is not helpful since a robot's appearance (passive cues) and behaviour (active signals) are created by roboticists for communicative purposes in a similar time frame. That is, the signals and cues are not necessarily ``evolved'' in the traditional sense. Moreover, distinguishing between the explicitness or implicitness in the \emph{design of signals} by the roboticist may be mistaken for the explicitness of \emph{the signal itself} (i.e. whether the signal needs interpretation by the receiver or not). In the suggested new typology, we address both these problems by focusing on communicative signals that might be intended or unintended and at the same time contain an explicit or implicit component.

\section{A new typology of communicative robot signals}
\label{sec:typology}

In this section, we introduce a new typology for communicative robot signals in \gls{hri} that aims to support roboticists in designing and evaluating robot behaviour.
Since the suggested typology heavily relies on ethological communication theories and draws inspiration from some earlier attempts at applying communication theories to \gls{hri}, we will first describe how the typology in general relates to existing theories and frameworks that consider communication as introduced in Sections~\ref{sec:ethology} and~\ref{sec:applying}.
We will then present how the typology looks in detail, proposing a selection of core properties to describe communicative signals and discuss how our suggested worksheet can be used to apply the typology to characterise an example behaviour. The remainder of the section will argue how the typology can guide new robotic behaviour design and the analysis of already conducted interaction studies.

\subsection{Scoping considerations}
\label{sec:typology:theory}

We adopt \citeauthor{Slater1985Ethology}'s view that communication is a behaviour that alters the behaviour of an interaction partner~\cite{Slater1985Ethology}. Contrarily to some definitions in ethology (see Sect.~\ref{sec:ethology:com-defs}), we neither require communicative signals to be evolved to be regarded as communicative nor do we require intentionality to consider a robot as a communicator. Moreover, we consider communication between robots and humans to be asymmetrical, similar to type II communication between different species~\cite{csanyi1988can}.

We want to stress that our notion of communicative signals also includes non-actions \cite{watzlawick1967pragmatics} like interruptions of robot movements, which sometimes happen deliberately, for example, to allow for human conversation, object recognition, or position estimation. Moreover, the typology also covers and directly addresses non-obvious (implicit) signals and overloaded signals carrying multiple messages.
We would further like to emphasise that we do not see communicative signals in isolation but rather embedded in an interaction~\cite{Holthaus2021encounter}, potentially involving multiple parties, and complex turn-taking behaviours and feedback channels, as described by \citet{DeVito1986Handbook}. However, to individually characterise communicative signals, we focus on a specific view considering a single robot as the sender and a single human as the receiver. This allows us to relate the typology to communication models introduced in Section~\ref{sec:applying:characterising}, e.g. \citeauthor{shannon1948mathematical} \cite{shannon1948mathematical}, while preserving the potential to consider multiple receivers or swapping roles.

While the proposed typology might be applicable to either the robot or the human as a sender, for simplicity here, we take a unidimensional perspective to discuss communicative signals in the context of particular robot behaviours. With that, we aim to provide a structured approach to interaction-centric robot behaviour design and analysis in light of the asymmetric nature of communication between humans and robots. We further acknowledge—but do not further discuss—that communicative signals can be affected by (external) noise, which might distort the message content.

\subsection{Characterisation of robot signals}
\label{sec:typology:signals}

\begin{figure}
    \centering
    \includegraphics[width=.69\linewidth,clip,trim={.63cm .63cm .63cm .63cm}]{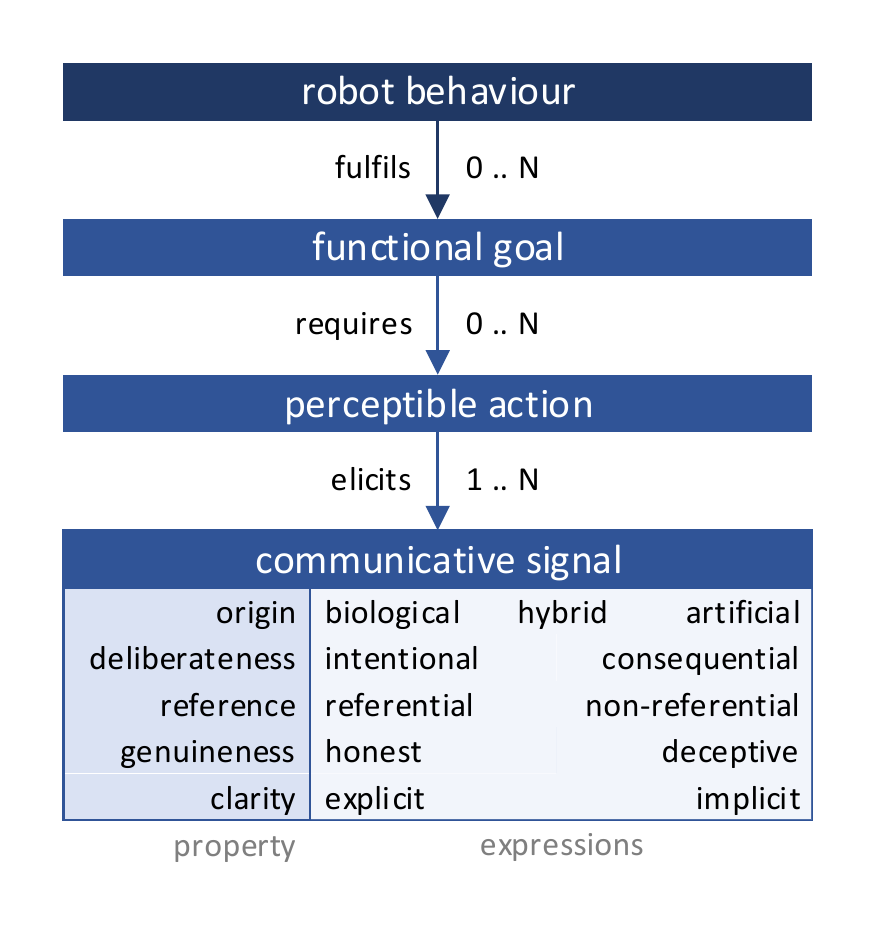}
    \caption{Typology of communicative robot signals in the context of robot behaviours where signals are elicited by perceptible actions required to fulfil a functional goal. Communicative signals can be characterised using five properties.}
    \label{fig:overview}
    \Description{This figure displays the typology as a directed graph with five nodes. The robot behaviour node points to a functional goal node, annotated with ``fulfils 0 to N'', which itself points to a perceptible action node in a ``requries 0 to N'' relation. The latter points to the communicative signal node, as ``elicits 1 to N''. Within the communicative signals node the five properties origin, deliberatenes, reference, genuineness, and clarity as well as their expressions are listed.}
\end{figure}

We define a robot behaviour as a higher level function aimed to fulfil a range of functional goals, which in turn might require certain robot actions that are perceptible by a human user (Fig.~\ref{fig:overview}). Each of the actions might have a number of components that the robot uses to reach the functional goal, potentially perceptible via a range of modalities. Such a robot action always elicits one or more communicative signals as long as a human user perceives it.

Like other animals, humans perceive communicative signals using multiple modalities including seeing, hearing, and touch~\cite{GoodrichHumanRobotInteraction2008,bonarini2020communication}. The typology focuses on structuring the content and properties of such non-verbal messages, i.e. describing what a signal might reveal about the robot  or the environment and how the signal might be interpreted by the human (Sect.~\ref{sec:applying:characterising}).
We therefore propose to individually characterise communicative robot signals in the context of \gls{hri}, addressing qualities of the sender and how it produces the signal, the signal's message content, and how much interpretation is required at the receiver's end.
While we focus on five core properties, the typology is meant to be extensible and welcomes others to define and discuss additional properties. Some of the propositions have proven to be useful when characterising human (or animal) signals, others are novel suggestions that arise from the specific nature of \gls{hri} where roboticists have control over the design and expression of all robot behaviour, including communicative signals.


\subsubsection{Origin} Designing social behaviours is complex with formal generation still requiring manual fine-tuning of signals~\cite{sequeira2016discovering}. However, solutions for manually signalling an intended meaning with a robot can often be derived from biology (Sect.~\ref{sec:ethology:robotics}). It might be worthwhile to consult ethology when assessing whether a signal is expressed correctly or can efficiently transport the intended meaning. On the other hand, imitating biology when artificial solutions are proven to be effective might lead to misunderstandings or increase design efforts unnecessarily. Hence, evaluating a signal from both perspectives allows roboticists to easily identify potential for improvement or reveal unintended side effects.

With the new typology, we thus suggest to determine the \property{origin} of a signal, i.e. whether the signal mimics communication between humans—or animals—or whether it has been designed otherwise to understand and model communication from the sender's (robot) perspective. We thereby encourage roboticists to reason about a signal's origin and explicitly think along both alternatives and what implications they entail with regard to signal design and evaluation.

Signals might have be derived from human communication but adapted to suit the needs of human-robot interaction, allowing this property to have three values: \expression{biological}, \expression{artificial}, or \expression{hybrid}.

\subsubsection{Deliberateness}

Consequential signals are a common source of misunderstandings \cite{Schulz2021MovementActs, mooreSoundImplicitInfluence2018, fridPerceptualEvaluationBlended2022, trovatoSoundSilenceInvestigating2018} because the designer cannot control what information is being sent and is often \emph{unaware} that information is being sent at all. We therefore propose to capture whether a signal has been produced deliberately or whether it is merely the accidental byproduct of an action to allow roboticists to identify and reduce unwanted effects in the interaction partner.
The identification of unintended signals might guide the design process of robots or the examination of unexpected user behaviours. One possible aim could be to eliminate unintended signals; another one would be to identify the signal's nature and use them deliberately.

Assessing the signal's \property{deliberateness} allows roboticists to further investigate a signal from the sender's perspective.
With this typology, we encourage roboticists to carefully consider whether perceptible actions only elicit intentional signals or whether they might result in a signal that is unintended. A single action may elicit more than one signal, possibly resulting in unintended secondary meanings, depending on robot state, context, and receiver.
This property can have two values: \expression{intentional} or \expression{consequential}.

\subsubsection{Reference}

Robot signals with a primarily social function are often designed to reveal some piece of information about the robot's inner state to its human users~\cite{fiore2013toward,Holthaus2021encounter,Esposito2016EmotionModel,Fischer2019Emotion}. We argue that assessing the motivational reference of a signal, similar to ethological approaches~\cite{Kampis1991Systems}, helps roboticists to determine whether the signal appropriately reveals or perhaps fails to reveal any of the robot's inner states. Moreover, by considering the reference of a signal, we aim to foster the identification of additional information that might potentially be carried by a signal. Referential signals like deictic gestures and functional behaviour, e.g. delays or pauses, also have a (possibly consequential) non-referential component that provides human users with information about the robot and shapes their expectations. Sophisticated voice or advanced arm movements when giving directions might suggest other parts of the robot, such as its perception modules, are advanced while in reality, they might not be.

When looking at the signal's message content we suggest determining a signal's \property{reference}, i.e. whether it considers information about the sender and its inner state or alternatively any outside entity. This allows us to distinguish between \expression{non-referential} and \expression{referential} communicative signals in a similar way that ethology does~\cite{Kampis1991Systems} for assessing the effectiveness of a signal.

\subsubsection{Genuineness}

Many communicative robot signals are deceptive in their nature~\cite{sharkey2021deception} and often suggest non-existing robot capabilities to a human. Displaying emotion, for example, dishonestly implies that a robot would be able to feel them. An analysis of this property might therefore help to evaluate positive and negative effects on the receiver, as well as potential harmful effects on the interaction like eliciting wrong assumptions or false expectations~\cite{rosen2021expectations} and generating inappropriate relationships between human and robot~\cite{ososky2014determinants,salem2015towards}.
Whether a signal is honest or deceptive is taken from the perspective of the robot, the information it has, and is independent of the \emph{truth} or facts of the interaction.
Missing information, failures in the accuracy of sensors, etc.\ changes whether the information communicated by the signal is true or false.
One could debate whether such a truth is possible in all situations, but that debate is beyond the scope of this paper.

Another important dimension of a communicative signal, relating mostly to its message content, is therefore its \property{genuineness}. Signals can either be \expression{honest} or \expression{deceptive}. Determining the genuineness of the signal can promote discussing whether a signal, perhaps intentionally, misleads the receiver or whether it communicates honestly and shapes user expectations as desired.

\subsubsection{Clarity}

Implicit signals are frequently applied to support interactive robots subtly transporting meaning, for example, to request help from a human in a collaborative scenario~\cite{cha2016collaboration} or by using social proximity to support interaction opening~\cite{Holthaus2021encounter}. Implicit signals account for the majority of communication in some situations, such as crossing intersections in traffic~\cite{Lee2020road,zhangAnalysisImplicitCommunication2022}, which stresses the need to identify and characterise them when considering autonomous vehicles as well as \gls{hri}. Implicitness of a signal can lead to ambiguity and therefore can provide flexibility and adaptability to the current situations, allowing humans to interpret the exact meaning of the signal (c.f.~\citep{pelikanAreYouSad2020}). On the other hand, such signals are often more difficult to decode than explicit signals~\cite{Smith2003AnimalSignals}, especially when considering the diversity of humans, coming from different backgrounds and bringing different expectations.

When considering the reception of the message, the typology therefore suggests to determine a signal's \property{clarity}. This property can have two values: \expression{explicit} or \expression{implicit}. That is, we aim to estimate whether the signal is sufficiently clear so that it can be understood without much effort or whether it is subtle, leaving room for or requiring interpretation by the receiver.

\subsection{Application of worksheet}
\label{sec:typology:examples}

\begin{figure*}
    \centering
    \includegraphics[width=.96\linewidth,clip,trim={.63cm .8cm .63cm .63cm}]{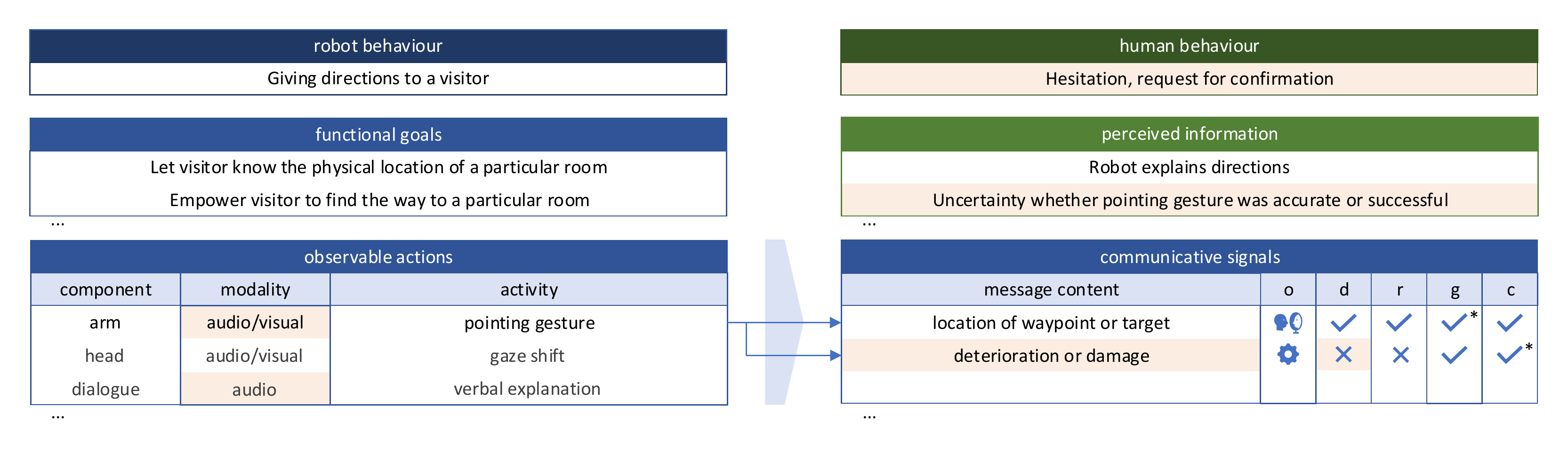}
    \caption{Recommended worksheet to use in robot behaviour design or analysis applied to an example behaviour where a receptionist robot gives directions to a visitor. Shaded entries denote areas where the attention of a roboticist might be required.}
    \label{fig:character-chart}
    \Description{This figure displays the worksheet as boxes where the nodes of Figure 1 are used as headings for boxes. Two boxes ``perceived information'' and ``human behaviour'' are added. Boxes are summarising the text in Section 4.3 where the "robot behaviour" box is filled with "Giving directions to a visitor", two functional goals and three observable actions are listed as a triplet consisting of component, modality, and activity (with the audio modality of the pointing gesture and verbal explanation shaded to indicate potential problems). The ``communicative signals'' box lists two signals with the message content ``location of waypoint or target'' and ``deterioration or damage'' resulting from the pointing gesture and displays checkmarks or crosses for the properties of each signal to indicate their expression, using symbols for biologically-inspired and artificial. Moreover, ``uncertainty'' is listed (shaded) as potentially problematic perceived information in the human, together with ``hesitation'' as a human behaviour.}
\end{figure*}

The typology of robot communicative signals is meant to be workable, extensible, and applicable to similar kinds of communication. For that, we introduce a worksheet (Fig.~\ref{fig:character-chart}) that may be used to apply the typology in the behaviour design and evaluation processes.

As an initial step for using the worksheet, we suggest to identify individual robot behaviours within a certain scenario, i.e. higher-level functions to fulfil a range of functional goals.
For each behaviour, the worksheet is then recommended to be filled left-to-right but please note that the filling order highly depends on the use case, for example, one might want to start on the right, denoting any human behaviour when analysing surprising events. We would like to discuss the worksheet alongside a simple example where a humanoid receptionist robot engages in a behaviour that explains directions to a visitor. We can therefore start filling the worksheet with capturing the \emph{robot behaviour} as such. The according \emph{functional goals} would be to communicate the room's physical location within the building and enable the visitor to find that room.

Then, the behaviour should be decomposed and partitioned into actions. This can be achieved by individually looking at each component involved in the behaviour and looking at the sequence of perceivable actions, including pauses, carried out by that component during the behaviour. Here, the direction-giving behaviour happens immediately after a human asks for a location. It is composed of a gaze shift moving the head and a pointing gesture with the arm that can be visually perceived. Both movements are also presented with a noise originating from the motors and moving parts. Simultaneously, the robot verbally explains where the room is located and what waypoints to find along the path.

All three actions are designed to multimodally contribute to the message content of explaining the location of waypoints or the final target. For simplicity, we continue the worksheet  characterising only the properties of the pointing gesture, omitting the gaze shift (functioning as intended), and the verbal explanation (implemented in a sub-optimal way, e.g. without lip synchronisation).

The pointing gesture would usually be modelled in a human-like (\expression{biological}) fashion, whereas the noise of moving mechanical parts or motors is \expression{artificial}. This would imply that the movement would likely be interpreted in comparison to human movement, highlighting the importance of eliciting a smooth trajectory that is biologically plausible. Jerky or unnatural movements on the other hand might suggest a faulty robot, potentially eliciting association with illness~\cite{Langer2020moving}.
A receptionist robot would generate a pointing gesture \expression{intentionally} to support its verbal explanation about the route. The presented noise, on the other hand, is often an unwanted \expression{consequence} of moving parts, likely causing problems in the interaction where unwanted signals are communicated to the human; for example, when exaggerated or distorted noise is produced due to insufficiently calibrated hardware.
Pointing gestures are inherently \expression{referential} whereas emitted noise solely informs about the robot's current condition and is hence \expression{non-referential}. A closer look at what the noise could communicate about the robot might help to explain unexpected human behaviour. Excessive noise, for example, might communicate deterioration or damage whereas the normal motor sound would appropriately signal the robot's inner workings and capabilities.
Since, in this example, the receptionist robot is designed to be helpful, it would produce an \expression{honest} gesture, pointing in the correct direction to the best of its knowledge. If desired, however, such a gesture could be generated to intentionally mislead a human user. Consequential noise is not created on purpose and therefore also usually honest.
Pointing gestures are often \expression{explicit}, especially when displayed while giving directions. However, if the sender and receiver used different reference systems, the gesture would become ambiguous and hence less explicit.
Likewise, a squeaking noise can explicitly signal that robot parts are grinding against each other whereas other noise might be more difficult to interpret and thus be more implicit.

Assessing the properties of the communicative signal allows us to estimate the information that is being perceived by the human user of the robot. In this example, the user notices that the robot explains directions, which is facilitated by all three actions. However, excessive noise while presenting the gesture might cause uncertainty about the capabilities of the robot and the accurateness of its actions. An expected (or observed) human behaviour may therefore include signs of hesitation or requests for confirmation.

\subsection{Using the typology in behaviour design}
\label{sec:typology:design}

Perceptible robot behaviour results from design choices at different levels of abstraction, ranging from low-level hardware design to the definition of high-level behavioural goals, which may involve a sequence of actions or a combination of actuators. At all levels, roboticists have the opportunity to influence the emerging robot behaviour so that it is, to some degree, shaped by a designer to perform a confined set of functions. Robot behaviour, however, is also dependent on perception and embedded in environment and interaction; not every robot action is predictable. For example, robot behaviour might be generated from a learning component adapting to user preferences. Still, the robot's interaction architecture, training pipelines for machine learning, and individual robot actions are frequently hand-crafted and can be dictated by a roboticist.

The proposed typology aims to be applicable at all abstraction levels of robotic behaviour design by looking at the final robot behaviour from an interactive perspective, i.e. by focusing on perceptible actions. At the same time, it is good practice to always consider the functional goals when designing any robotic behaviour~\cite{miklosiUtilizationSocialAnimals2012}. As described in Sect.~\ref{sec:typology:examples}, the worksheet supports such an approach since it encourages roboticists to associate each observable robot action aiming to fulfil a functional goal as part of a high-level behaviour with communicative signals. When designing behaviours, signals and their properties emerging from observable actions can then be characterised in context using the suggested terminology, which in turn allows estimating what information is transmitted and how the human's behaviour might be affected.

Someone might want to design an \textit{attentive behaviour} for a humanoid robot that aims to fulfil the functional goals of \textit{attracting users} and \textit{keeping them engaged} during conversation. For that purpose, the robot could be equipped with a function that makes it \textit{look towards a human}, letting the user observe robot actions where it continuously adjusts its \textit{body} and \textit{head} orientation as well as its \textit{eye} gaze direction towards the user and the user's face. For the purpose of this exercise, we assume that the only involved modality is vision whereas the motors operate without perceptible noise.

Such activities send many communicative signals a roboticist might want to control during the design process. All three actions can be designed to contribute to a \expression{biologically}-inspired, \expression{intentional}, \expression{non-referential}, \expression{honest}, and largely \expression{explicit} communicative signal indicating that the robot's attention is focused on the human. However, a designer may want to reinforce the signal by adding another \expression{artificial} one, such as setting the robot's eye colour to green whenever it has detected a user in its focus. At the same time, changes in body posture and gaze can imply \expression{consequential} communicative signals that may need to be considered. Over-commitment to a focused interaction might be such a signal, for example when orienting the body towards a human who is still far away or by perpetually staring at them during conversation. Moreover, signalling attention while the robot is (temporarily) unable to interact might set wrong expectations and therefore become \expression{deceptive}. A robot designer could use such insights (and more) to adjust the situations where the behaviour is being expressed to maintain the desired human behaviour of being engaged with the robot.

\subsection{Using the typology in existing studies}
\label{sec:typology:apply}

While we have provided examples of a robot pointing and adjusting its attention, let us apply the topology to some instances from literature.

The first example is the Fetch robot resetting its navigation stack by rotating in a circle (described in \citet{Schulz2021MovementActs}). The origin of this action is \expression{artificial} and is not based on anything biological.
Looking at the rotations deliberateness, we can classify the signal as \expression{consequential} since the robot's motion was to reset its navigation stack than communicate anything (but it could be interpreted by an observer as intentional).
The rotation itself was \expression{non-referential} since the robot's motion is related to  its internal state of being unable to localise itself, not referencing the environment.
As to the rotation's genuineness, we can classify it as \expression{honest}; The robot was trying to reset its navigation stack and not being deceptive.
As to the rotation's clarity, it was clearly \expression{implicit}.
People observing the robot would have no idea what it was doing unless they were familiar with the robot's navigation routine.
Applying the new typology shows there is potential for this motion to be misunderstood even though the behaviour serves its function.
It seems that the implicitness of the signal makes people confuse the non-referential repair mechanism with a referential signal that aims to draw their attention to something in the robot's
surroundings.

For the Cozmo robot signal that it did not get a fist bump in time (described in \citet{pelikanAreYouSad2020}), the origin of the action is \expression{biological} (human).
Here, the deliberateness of the action was \expression{intentional}, as it signalled that the time had expired for giving a fist bump.
This is a \expression{non-referential} signal as well since Cozmo signalled its ``desire'' for a fist bump on its front loader.
The signal was also \expression{honest}, as a desire for this action has been modelled on the robot.
The clarity, at least as thought by the designers, was explicit.
The typology shows that the understanding of the signal should be straightforward, yet the behaviour in isolation could still be misinterpreted~\citep{pelikanAreYouSad2020}.

The typology can also be used to see differences in behaviour between conditions in a study (and possibly improve the internal validity of an experiment).
For example, in the study of the robot waiting for an elevator from \citet{galloExploringMachinelikeBehaviors2022}, the origins of the ``machine-like'' behaviour is \expression{artificial}, and the ``human-like'' behaviour origin is \expression{biological}, but the other properties of both behaviours are \expression{intentional}, \expression{explicit}, \expression{honest}, and \expression{referential}. Using the typography can show how changing one component of the communicative signal can have an effect on the behaviour participants prefer in a given situation. 

\section{Discussion and Limitations}
\label{sec:discussion}


The typology provides an accessible starting point for designing behaviours or examining how robot behaviour could be interpreted. The typology should work for robot signals in other modalities such as virtual or augmented reality~\citep{walkerCommunicatingRobotMotion2018}. It also has the advantage that it can be compatible with other theories and explanations while at the same time being extensible to allow for customisation.
For someone who wanted to use movement acts as a way to explain movement, the typology provides a mechanism for breaking down the act into implicit or explicit and intentional or consequential parts, while also providing opportunities for adding richness in describing the movement act (such as deciding if the act is honest, referential, etc.).
Designing signals and behaviours using the new typology can be incorporated into other theories of communication, such as the AMODAL-HRI~\cite{frijns2021communication}. In that model, our typology can provide a starting point to discuss how the robot's signals that can lead to joint actions with humans. This matches suggestions that ``Analysis at the level of signals and cues can serve as a basis for discussion, but needs to go further.''~\citep[][p.~11]{frijns2021communication}. 

\citeauthor{miklosiUtilizationSocialAnimals2012} posited: ``… the actual social behaviour of a robotic companion should depend on the assumed function'' ~\citep[][p.~2]{miklosiUtilizationSocialAnimals2012}. The typology reinforces this importance of behaviour design for a functional goal. The worksheet (Fig.~\ref{fig:character-chart}) supports this process and facilitates relating the functional goals with the communicative meaning of signals and the predicted or observed human behaviour.

The clarity of a robot signal depends on any expectations that a human might have, which might originate from the robot's intended function or perhaps the human's unfamiliarity with the robot. User expectations can sometimes determine whether certain signals have a positive effect on user acceptance and the robot's trustworthiness. Some people who interact with a humanoid robot, might expect and prefer a robot to display emotions, which then in turn increases the robot's acceptability and trustworthiness~\cite{Esposito2016EmotionModel} even when the emotion is clearly deceptive. Human expectations thereby vary with a magnitude of factors, which is not directly addressed in our typology. However, \citeauthor{rosen2022social} provide a framework~\cite{rosen2022social} that can be considered when applying the typology in behaviour design or analysis to estimate the effects of user expectations on the signal, the perceived information and the human behaviour.

\citet{pelikanAreYouSad2020} recommended designers explore the ambiguity of implicit signals to enhance interaction.
The new typology is limited in assessing an exact level of implicitness, but it can provide a good place to see this ambiguity by requiring designers to classify the clarity of the signals.
We therefore encourage designers to classify the clarity in a binary way, however, comments or more gradual classifications might also be applied. Moreover, diversifying participants and other stakeholders can provide opportunities for challenging assumptions and finding solutions.

Outside the scope of the typology are any costs of the robot producing communicative signals that need to be weighed against any potential benefit of communicative success. In the animal world, producing a communicative signal often means balancing between the risk of becoming prey versus increasing the individual's chance of reproduction~\cite{Smith2003AnimalSignals}. Human-human or human-robot communication typically has much lower predatory risks, but other costs might become relevant, e.g. when balancing attention. Communicative robot gaze towards a person, for example, might distract from an object classification task while providing the benefit of easily establishing joint attention.
Finally, we have shown possible applications for the typology as an initial verification. Future verification can include \gls{hri} experiments that examine behaviour design and communication with or without the help of the typology.


\section{Conclusion}
\label{sec:conclusion}

In this article, we examined how ethology could be applied to
\gls{hri}, looked at example studies that used ideas from ethology or
other areas that have a robot communicate with humans, and examined
different ways of describing communication between them. We then
presented a new typology for communicative robot signals that included multiple properties of origin, deliberateness, reference, genuineness, and
clarity. We used this typology to examine communication from previous
studies and provided a sketch to how typology could be used for new
designs.

Typologies and theories may be refined and new ones will be
proposed; we complement this typology with a worksheet that can be applied to encourage designers and roboticists not only to think about a
behaviour, but also assess how behaviours will be perceived and
understood by people interacting with robots. Any increased focus
on this aspect of \gls{hri} should result in less confusion and more
understandable interactions. This may lead to future
interactions that are more effective, acceptable, trustworthy, and pleasant.

\bibliographystyle{ACM-Reference-Format}
\bibliography{bibliography}


\end{document}